\documentclass[journal]{IEEEtran}
\ifCLASSINFOpdf
  \usepackage[pdftex]{graphicx}
  \graphicspath{{photos/}}
  \DeclareGraphicsExtensions{.pdf,.jpg,.eps}
\else
\fi
\ifCLASSOPTIONcompsoc
  \usepackage[caption=false,font=normalsize,labelfont=sf,textfont=sf]{subfig}
\else
  \usepackage[caption=false,font=footnotesize]{subfig}
\fi
\usepackage{graphicx}
\usepackage{amsmath}
\usepackage{natbib}
\usepackage{multirow}
\setcitestyle{numbers,square}

\begin{document}
%
\title{Encoder-Decoder based CNN and Fully Connected CRFs for Remote Sensed Image Segmentation}
%
%
%

\author{Vikas Agaradahalli Gurumurthy,~\IEEEmembership{Independent Researcher}
\thanks{Author e-mail: ag.vikas98@gmail.com.}
}

\maketitle

\begin{abstract}
With the advancement of remote-sensed imaging large volumes of very high resolution land cover images can now be obtained. Automation of object recognition in these 2D images, however, is still a key issue. High intra-class variance and low inter-class variance in Very High Resolution (VHR) images hamper the accuracy of prediction in object recognition tasks. Most successful techniques in various computer vision tasks recently are based on deep supervised learning. In this work, a deep Convolutional Neural Network (CNN) based on symmetric encoder-decoder architecture with skip connections is employed for the 2D semantic segmentation of most common land cover object classes - impervious surface, buildings, low vegetation, trees and cars. Atrous convolutions are employed to have large receptive field in the proposed CNN model. Further, the CNN outputs are post-processed using Fully Connected Conditional Random Field (FCRF) model to refine the CNN pixel label predictions. The proposed CNN-FCRF model achieves an overall accuracy of 90.5\% on the ISPRS Vaihingen Dataset. 
\end{abstract}


%
\IEEEpeerreviewmaketitle

\section{Introduction}
%
%
%
%
\IEEEPARstart{A}{n} important intermediate step in the automation of map generation and updation from raw images is semantic segmentation. Semantic segmentation is the task of assigning each pixel in input image with the class label it is likely to belong to. Though numerous works in the past decades have contributed to improving image segmentation techniques, complete automation of map generation is yet to be achieved. In an urban scene, disparate objects with similar visual/spectral signatures and homogeneous objects with varied visual/spectral signatures pose a challenge to segmentation algorithms.
	
Since their advent, the convolutional neural networks (CNN) have become the benchmark in computer vision tasks. They have outdone the traditional methods in tasks such as regression, classification, object detection and semantic segmentation. AlexNet \cite{NIPS2012_4824}, a CNN based model for image classification, won the ImageNet challenge in 2012. \cite{long2014} converted a CNN trained for classification to a fully convolutional neural network (FCN) that could be trained end-to-end pixel-to-pixel for semantic segmentation. Their FCN model achieved state-of-the-art performance on PASCAL VOC, NYUDv2, SIFT Flow datasets. Several works have adopted the supervised learning approach based on CNNs for analysing remote sensed images. \cite{1d2d} compared ensemble of 1D-CNNs with convolutions in spectral domain and ensemble of 2D-CNNs with convolutions in spatial domain to obtain pixel-by-pixel prediction of class labels. Their work concluded ensemble of 2D-CNNs to be superior. \cite{kemker2018algorithms} used more efficient FCN based models - sharpmask \cite{sharpmask} and refinenet \cite{refinenet} - to benchmark their multispectral dataset (RIT-18). The models improvise on standard skip architectures to merge features from shallow layers. Skip connections are known to facilitate gradient propagation across long-range connections and refine class boundaries in segmentation output. They observed that pre-training the models with synthetic data prior to training with actual data increased performance. \cite{marmanis} used an extension of architecture in \cite{long2014}. They used two different pathways with same architecture for image and digital elevation model (DEM) data, and, merged the spectral and height features shortly before the final layer that outputs the class probabilities. Predictions were averaged from several trained models of same architecture with different initialization and fully connected conditional random fields were employed for post-processing. 

Provided high intra-class variance and low inter-class variance in VHR imagery, it is intuitive to have large receptive field to incorporate features from a larger context rather than small local context for segmentation. Increasing the number of convolution layers or size of convolutional filters are among well known methods to increase the receptive field. \cite{Chen2014SemanticIS} introduced dilated/atrous convolution for time efficient image segmentation. Atrous convolutions expand receptive field by using gaps in convolution filters while keeping the computational budget constant. \cite{erf} introduced the concept of effective receptive field and showed atrous convolution to increase effective receptive field. In this work, atrous convolutions are adopted in the CNN model to have a large receptive field. The proposed atrous convolutions based model has symmetric encoder-decoder architecture with skip connections. Upsampling in decoder is achieved using transpose convolutions with overlapping stride.

\begin{figure*}
  \centering
    \includegraphics[width=\textwidth, height=175pt]{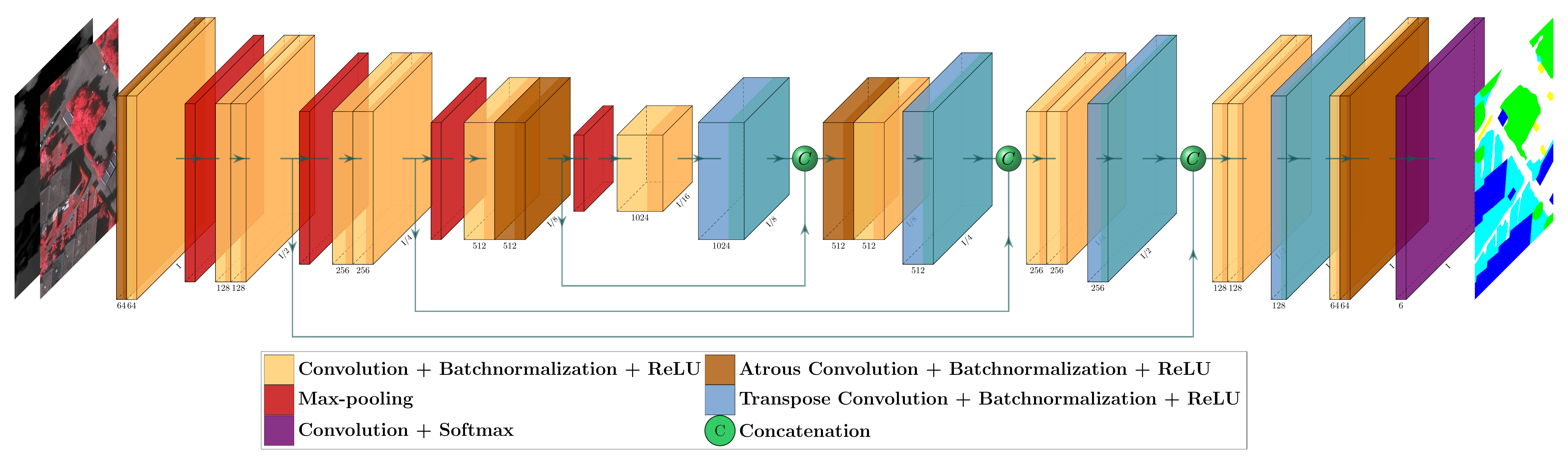}
    \caption{Architecture of the symmetric encoder-decoder model with atrous convolutions.}
    \label{fig:architecture}
\end{figure*}


Conditional Random Fields (CRFs) are popularly employed as post-processing step to smooth noisy segmentation outputs. \cite{Paisitkriangkrai} used edge sensitive binary CRF to refine segmentation results from a CNN. \cite{fcrf} proposed an efficient approximate inference algorithm for Fully Connected Conditional Random Fields (FCRF). Their results demonstrate that accounting for long range dependencies with dense pixel-level connectivity significantly improves segmentation accuracy. In this work, the FRCF algorithm proposed by \cite{fcrf} is integrated on top of the CNN model. Atrous convolution based model with FCRF post-processing provides a competitive overall accuracy of 90.5\% on the ISPRS 2D semantic labeling Vaihingen dataset.

The remainder of the paper is organized into following sections. Section \ref{sec:dataset} describes the dataset. The models considered are detailed in section \ref{sec:models}. The experimentation details and results are provided in section \ref{sec:exp}. Conclusion is drawn in section \ref{sec:conclusion}.

\section{Dataset}
\label{sec:dataset}

The Vaihingen 2D dataset provided by semantic labeling contest\footnote{\label{1}http://www2.isprs.org/commissions/comm3/wg4/semantic-labeling.html} of the ISPRS WG III/4 is utilized in this work for validation of proposed methods. Vaihingen is a small village in Germany with many detached and small multi-storied buildings. The dataset consists of 33 image tiles extracted from a large orthomosaic image with a ground sampling distance of 9cm. The images are provided as 8-bit TIFF files with three spectral bands - near infrared, red, green (IRRG). The label images classified manually at pixel level are provided as 8-bit TIFF files. Pixels are categorized into one of the following classes - impervious surface, building, low vegetation, tree, car, background. In addition to images and label images, the normalized Digital Surface Model (nDSM) images provided by \cite{gerke} are used for training the CNN models. The nDSM images are provided as 8-bit JPEG files.

\section{Models for Segmentation}
\label{sec:models}

\subsection{CNN model with atrous convolutions}

The proposed CNN model with atrous convolution operations is depicted in the Figure \ref{fig:architecture}. It has an encoder-decoder architecture with skip connections. The description of colored blocks is provided in the Figure \ref{fig:architecture}. The model has 4 atrous convolution blocks included as depicted with the rationale of increasing receptive field. Each convolution operation is followed by batch-normalization and non-liner activation using ReLU function. The terminating convolution is followed by softmax activation. Filters of size 3$\times$3 are used for convolutions and atrous convolutions. A dilation rate of 2 is used for atrous convolutions. The terminating convolution operation uses 6 (number of classes) filters of size 1$\times$1. Transpose convolutions are employed to undo the downsampling by maxpooling operations. They use filters of size 5$\times$5 with stride 2 and are followed by batch-normalization and ReLU activation. Skip connections are included to merge feature maps from encoder convolutions with feature maps of corresponding dimensions output from transpose convolutions. This model is referred as atrous convolution model in rest of the paper.

The results of segmentation from atrous convolution model are compared with those from a deep CNN model that lacks atrous convolutions. The model is referred to as standard convolution model. In addition, filters of size 2$\times$2 with stride 2 are used for transpose convolutions in the standard convolution model. No batch-normalization or non-linear activation follow transpose convolutions. Except for these differences the architecture of standard convolution model is same as that of atrous convolution model.  

\subsection{Fully connected conditional random field}
\label{sec:fcrf}
The standard convolution and atrous convolution models are integrated with fully connected conditional random field (FCRF) model\footnote{\label{2}https://github.com/lucasb-eyer/pydensecrf} proposed by \cite{fcrf} for post-processing. The model energy function is given by Equation \ref{eq:gibbs_energy}. 

\begin{equation}
\label{eq:gibbs_energy}
E(x) = \sum_{i} \theta_u (x_i) + \sum_{ij} \theta_p (x_i, x_j)
\end{equation}

$\theta_u$ is the unary potential evaluated as the negative logarithm of softmax probabilities ($- log(P_{x_i})$). The pair wise potential $\theta_p$ is $\mu (x_i, x_j) \sum_{m=1}^{K} w_m k^m (f_i, f_j)$, where $k^m$ are the guassian kernels which depend on feature vectors ($f$) for pixels i and j in arbitrary feature space, $w_m$ are weight parameters and $\mu$ is potts compatibility function. The kernels, based on position ($p$) and color ($I$) terms, adopted in the model are defined in Equation \ref{eq:kernels}

\begin{equation}
\label{eq:kernels}
\begin{split}
k(f_i, f_j) &=\hspace{3pt} w_1  \hspace{4pt} exp \left(- \frac{||p_i - p_j||^2}{2\sigma_{\alpha}^2} - \frac{||I_i - I_j||^2}{2\sigma_{\beta}^2} \right) \\ 
&+\hspace{3pt} w_2 \hspace{4pt} exp \left(- \frac{||p_i - p_j||^2}{2\sigma_{\gamma}^2} \right)
\end{split}
\end{equation}

$\sigma_{\alpha}$, $\sigma_{\beta}$, $\sigma_{\gamma}$ are standard deviations of the gaussian kernels. The message passing step under a mean field approximation to the CRF distribution can be expressed as gaussian filtering in feature space. Efficient high-dimensional filtering algorithms reduce the complexity of message passing resulting in an approximate inference algorithm that is significantly fast. For further details, readers are referred to the original paper \cite{fcrf}.

\section{Experimentation}
\label{sec:exp}
The experiment is carried out with standard convolution and atrous convolution models along with their FCRF integrated variants for post processing. All models are trained using training split of the ISPRS 2D semantic labeling contest Vaihingen dataset. For evaluation of the trained models the test split of the dataset is used. The f1-score and overall accuracy metrics obtained using accumulated confusion matrix are used to evaluate the performance. 

\subsection{Training}

The image tiles in the dataset labeled 1, 3, 5, 7, 11, 13, 15, 17, 21, 23, 26, 28, 30, 32, 34, 37 were used for training the CNN models. Each image tile is a true orthophoto (TOP) extracted from a large true orthomosaic photo. The image tiles were cropped into patches and subjected to augmentation due to computational bottleneck and to enable use of relatively large mini-bacth size. This resulted in a dataset consisting of 16244 patches of size 128 $\times$ 128. The patches were split into training and validation sets randomly in the ratio 3:1 during training. The input to the CNN models was IRRG image concatenated with nDSM along the channel dimension. The softmax output from model along with one-hot encoded labels was used to calculate the weighted cross entropy loss (Equation \ref{cr_loss}). 

\begin{equation}
\label{cr_loss}
l_{wce} = - \sum_{ij} w y_{ij} \log (y_{ij}^{p})
\end{equation}  

where $l_{wce}$ is the weighted cross entropy loss, $w$ is the array of weights associated with each class, $y_{ij}$ is the one-hot encoded class labels for pixel at $(i, j)$ and $y_{ij}^{p}$ is the softmax probabilities obtained from CNN model. Backpropagation and adam optimizer were used to update the parameters during training. A mini-batch size of 16 with a learning rate of 1e-4 was used. Weights ($w$) associated with background, building, car, impervious surface, low vegetation and tree classes are [5, 1, 100, 1, 2, 1]. The choice of weights is driven by class imbalance and common misclassifications.   

\begin{figure}
  \centering
    \includegraphics[trim={0 6.5cm 3cm 0}, clip, width=250pt, height=110pt]{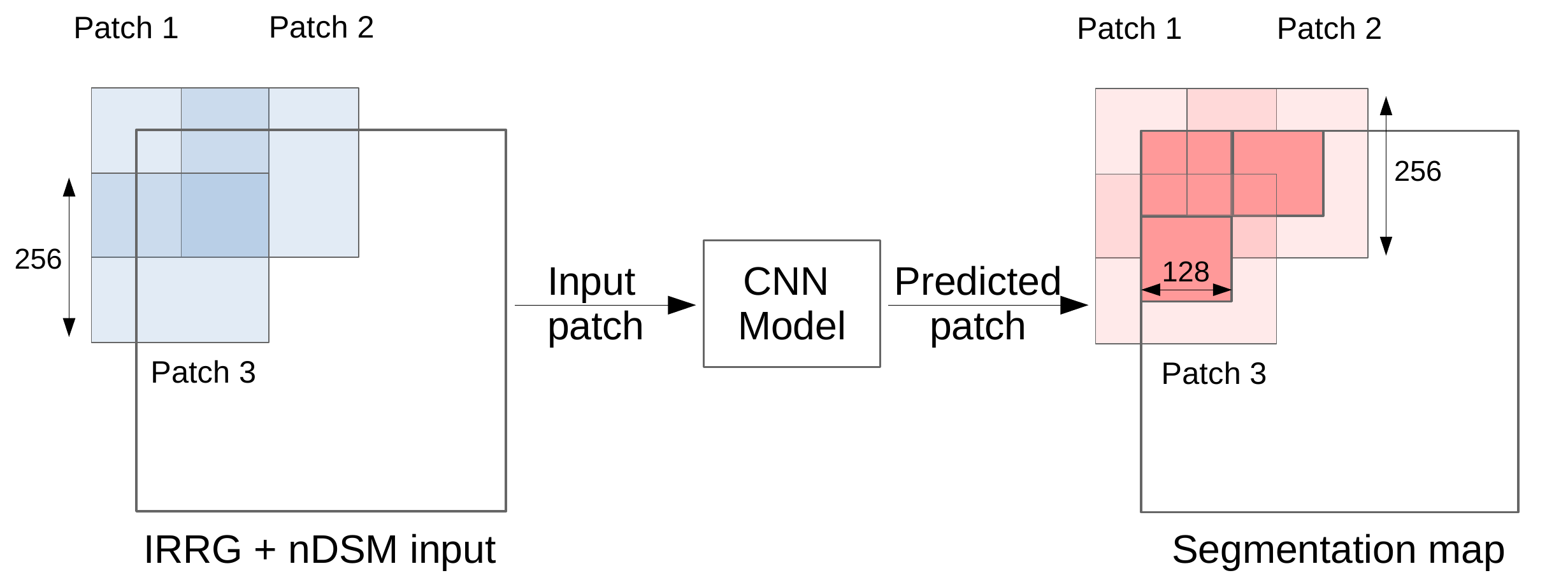}
    \caption{Patchwise prediction scheme}
    \label{fig:pp}
\end{figure}

\subsection{Evaluation}

The standard convolution (SC) and atrous convolution (AC) models, and their FCRF integrated variants (SC-FCRF and AC-FCRF) are evaluated using the test split provided by the dataset. The image tiles labeled 2, 4, 6, 8, 10, 12, 14, 16, 20, 22, 24, 27, 29, 31, 33, 35, 38 in the dataset form the testing set. During testing, the images are converted to patches of size 256 $\times$ 256 with 50\% overlap with adjacent patch. The central 128 $\times$ 128 region in the predicted 256 $\times$ 256 output from CNN model is considered. Patch-wise predictions are put together to obtain segmentation map with size equal to input size. Illustration of patch wise prediction scheme is presented in Fig. \ref{fig:pp}. The full segmentation maps from both CNN models are processed using FCRF model discussed in Section \ref{sec:fcrf}. Post-processing smoothens segmentation map by removing isolated noisy regions and improves class boundaries. 

The performance of the CNN models and their FCRF integrated variants is measured using f1-score and overall accuracy metrics. The definition of metrics is provided in Equation \ref{eq:metrics}. Confusion matrix is obtained for segmentation map of each test image tile. The confusion matrix contains references along row direction and predictions along column direction. True Positive (TP) pixels are obtained from the principal diagonal elements, False Positive (FP) pixels are evaluated as sum per column excluding principal diagonal element and False Negative (FN) are evaluated as sum per row excluding principal diagonal element. The metrics are evaluated using accumulated confusion matrix obtained by summing confusion matrices for individual image tiles. The overall accuracy reported is the fraction of trace and sum of elements of accumulated confusion matrix. The evaluation is carried out using labels with eroded boundaries provided in the dataset. 
\\[-2ex]
\begin{equation}
\label{eq:metrics}
\begin{split}
&Precision (P) = \frac{TP}{TP + FP}\\
&Recall (R) = \frac{TP}{TP + FN}\\
&F1-score = \frac{2*P*R}{P + R}\\
&Accuracy = \frac{trace(Conf. \hspace{3pt} matrix)}{sum(Conf. \hspace{3pt} matrix)}\\
\end{split}
\end{equation}

\subsection{Results}

\begin{table*}[ht!]
\caption{F1-scores and overall accuracies}
\label{tab:f1oa}
\centering
\begin{tabular}{|l|c|c|c|c|c|c|} 
\hline
\multirow{2}{*}{\bfseries Model} & \multicolumn{5}{c|}{\bfseries F1-score} & \multirow{2}{*}{\bfseries Overall accuracy}\\
\cline{2-6}
& \textbf{Building} & \textbf{Car} & \textbf{Imp. surf.} & \textbf{Low veg.} & \textbf{Tree} & \\ \hline
SC       & 94.2          & \textbf{83.5} & 90.7 & 81.6 & 88.2 & 88.9 \\ \hline
AC       & 94.8          & 83.1          & 91.5 & 83.1 & 88.9 & 89.8 \\ \hline
SC-FCRF  & 94.9          & 82.3          & 91.6 & 83.6 & 89.1 & 89.9 \\ \hline
AC-FCRF  & \textbf{95.3} & 81.5          & \textbf{92.2} & \textbf{84.6} & \textbf{89.6} & \textbf{90.5} \\ \hline
\end{tabular}
\end{table*} 

Classwise F1-scores and overall accuracies for the models considered are tabulated in Tab. \ref{tab:f1oa}. Performance evaluation metrics clearly indicate the AC model to provide increased performance over SC model. Post-processing segmentation maps with FCRF to remove isolated noisy regions and misclassifications, and refine segmentation boundaries has shown to further increase the overall accuracy. SC-FCRF and AC models provide similarly good results. AC-FCRF model delivers best results with highest f1-score for most classes and highest overall accuracy. Accumulated confusion matrix, normalized with respect to reference, for the test image tiles obtained using best performing model is presented in Tab. \ref{tab:acm}.

\begin{table}
\caption{Normalized accumulated confusion matrix for the test set using AC-FCRF model}
\label{tab:acm}
\centering
\begin{tabular}{|l|c|c|c|c|c|}
\hline
\textbf{Building}   & \textbf{95.56}	& 0.02	& 3.26	& 0.84	& 0.31 \\ \hline
\textbf{Car}        & 7.44	& \textbf{73.14}	& 18.57	& 0.43	& 0.29 \\ \hline
\textbf{Imp. surf.} & 2.37	& 0.14	& \textbf{93.76}	& 2.92	& 0.81 \\ \hline
\textbf{Low veg.}   & 1.71	& 0	    & 4.61	& \textbf{83.89}	& 9.79 \\ \hline
\textbf{Tree}       & 0.33	& 0	    & 1.17	& 8.84	& \textbf{89.65} \\ \hline
\end{tabular}
\end{table}  

\begin{figure}[!h]
\captionsetup[subfigure]{labelformat=empty}
\subfloat[]{\raisebox{0.18\linewidth}{\rotatebox[origin=t]{90}{IRRG}} \includegraphics[width=0.45\linewidth, height=0.4\linewidth]{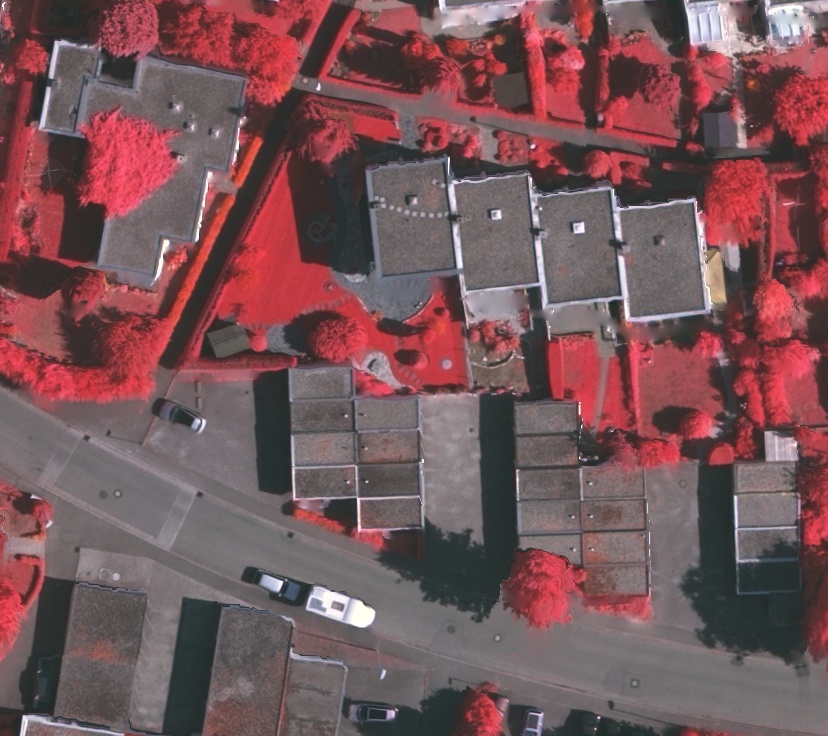}}
\hfil
\subfloat[]{\includegraphics[width=0.45\linewidth, height=0.4\linewidth]{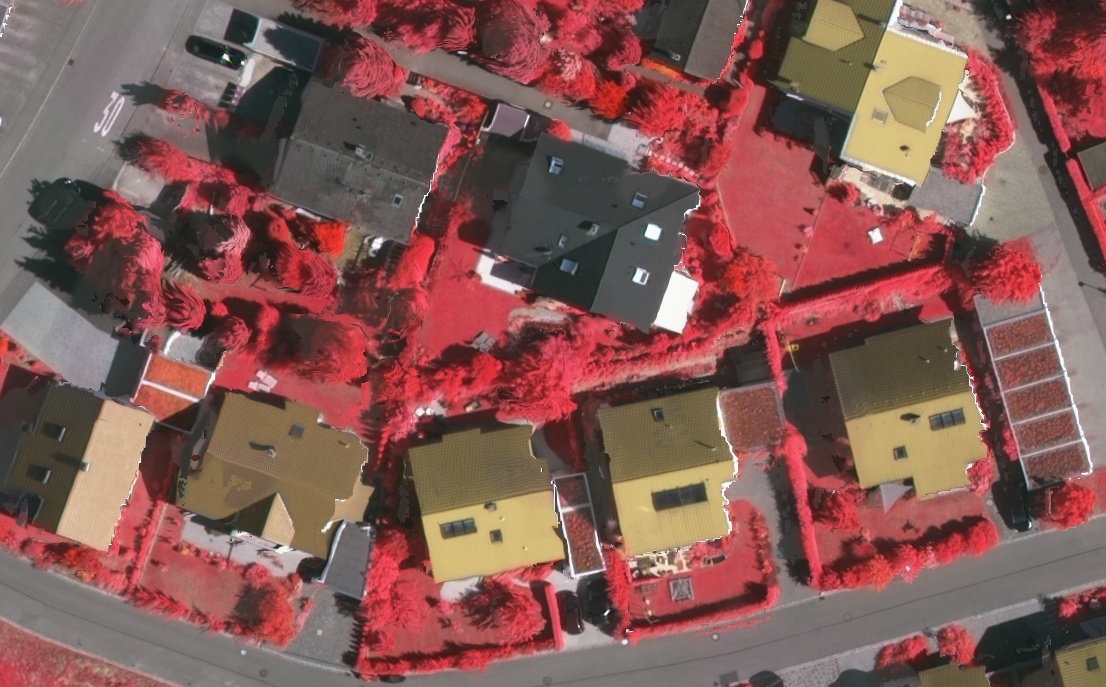}} \\[-3.5ex]
\subfloat[]{\raisebox{0.18\linewidth}{\rotatebox[origin=t]{90}{Labels}} \includegraphics[width=0.45\linewidth, height=0.4\linewidth]{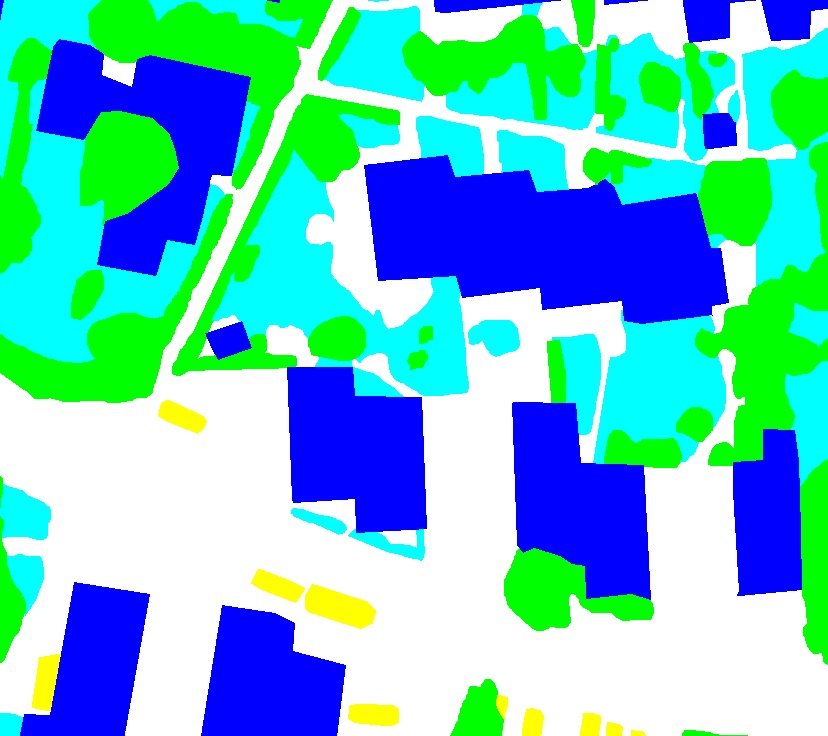}}
\hfil
\subfloat[]{\includegraphics[width=0.45\linewidth, height=0.4\linewidth]{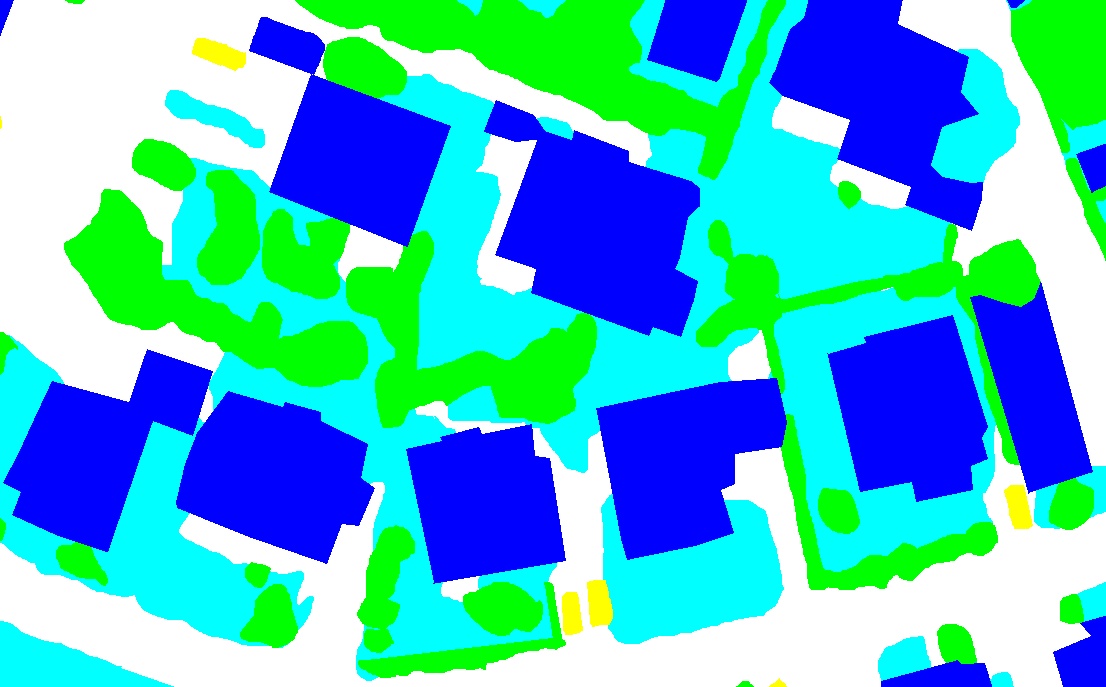}} \\[-3.5ex]
\subfloat[]{\raisebox{0.18\linewidth}{\rotatebox[origin=t]{90}{AC}} \includegraphics[width=0.45\linewidth, height=0.4\linewidth]{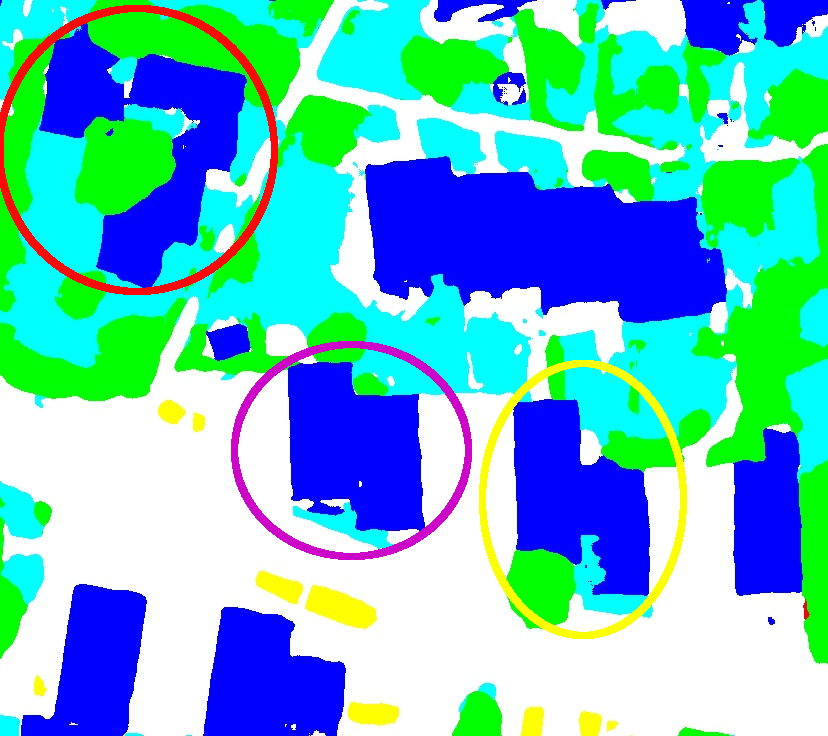}}
\hfil
\subfloat[]{\includegraphics[width=0.45\linewidth, height=0.4\linewidth]{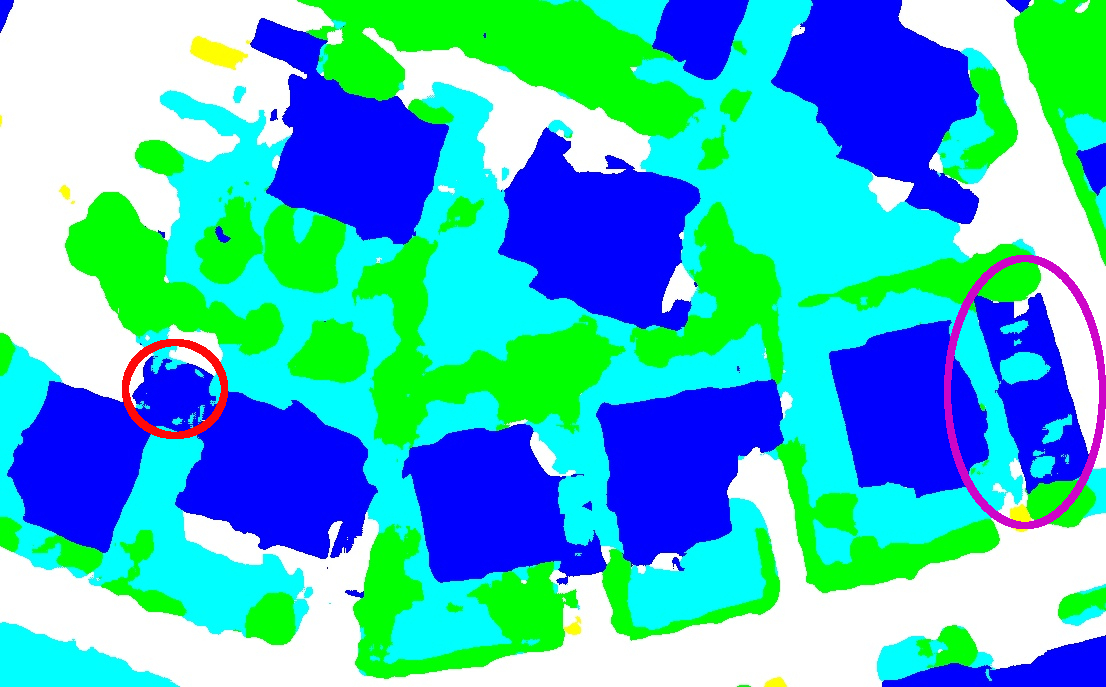}} \\[-3.5ex]
\subfloat[]{\raisebox{0.18\linewidth}{\rotatebox[origin=t]{90}{AC-FCRF}} \includegraphics[width=0.45\linewidth, height=0.4\linewidth]{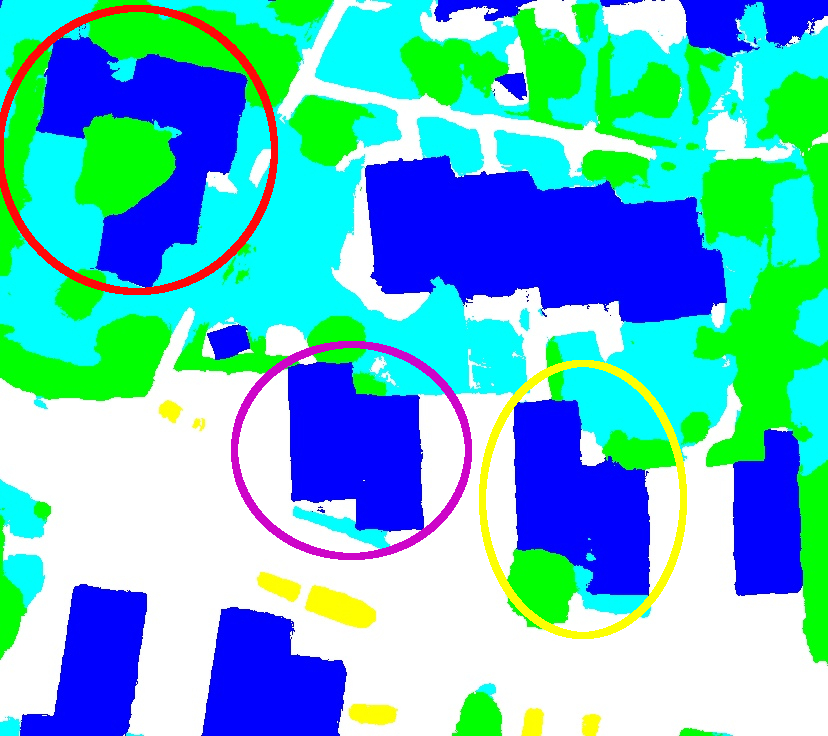}}
\hfil
\subfloat[]{\includegraphics[width=0.45\linewidth, height=0.4\linewidth]{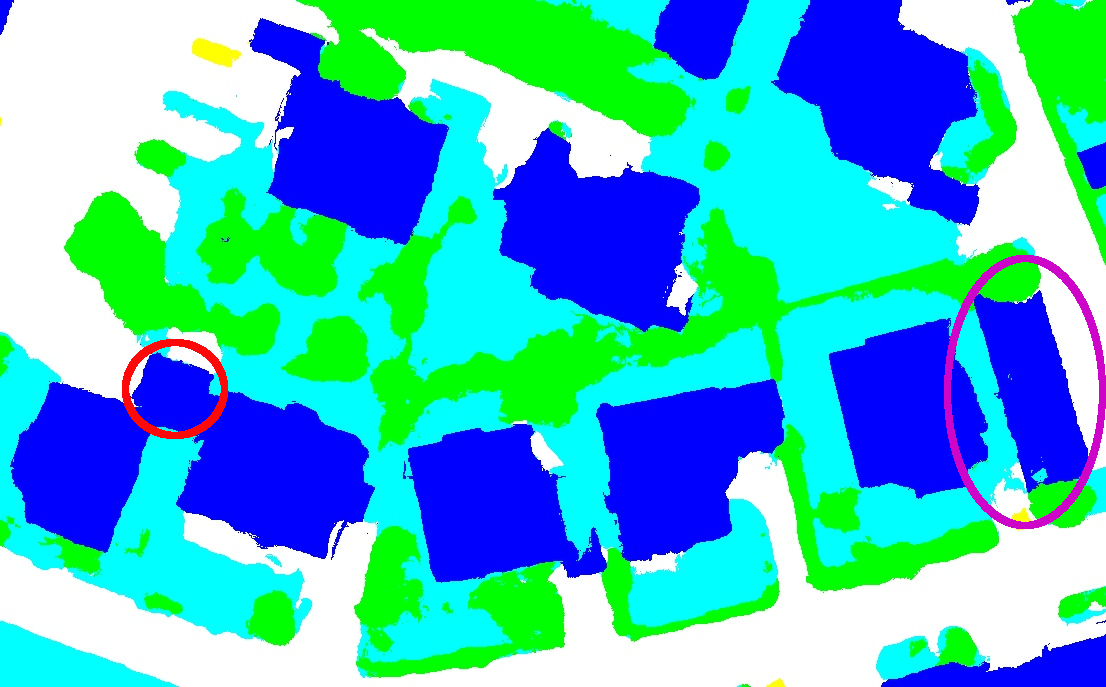}} \\[-3.5ex]
\caption{Comparison of segmentation output from atrous convolution model (AC) and its FCRF integrated variant (AC-FCRF).}
\label{fig:seg_comp}
\end{figure}

Visualization of segmentation maps from AC and AC-FCRF models is provided in Fig. \ref{fig:seg_comp}. The AC-FCRF model provides an increased overall accuracy of 0.7\% over AC model. Regions in segmentation maps highlighted with colored circles show the improvement in prediction caused due to post-processing using FCRF model. The boundary of buildings is refined and some misclassified pixels have been corrected in segmentation map of the first image. Isolated noisy regions within segmented buildings are removed in segmentation map of the second image. It can be seen from Fig. \ref{fig:seg_comp} and Tab. \ref{tab:f1oa} that FCRF model improves the results both visually and quantitatively.

Employing large receptive field has had a positive impact on performance in \cite{large_rf} and \cite{multiscale}. AC model achieves higher performance than the SC model. The key elements in AC model being atrous/dilated convolutions, and transpose convolutions with large filter and overlapping stride. Increasing receptive field has empirically shown to increase the prediction accuracy on a dataset with high intra-class variance and low inter-class variance. Further, post-processing the segmentation maps from AC model has provided an overall accuracy of 90.5\%. The results obtained using AC-FCRF model on ISPRS 2D Vaihingen dataset are competitive. 

\section{Conclusion}
\label{sec:conclusion}
In this paper, a deep CNN model is proposed for semantic segmentation of remote sensed images. The proposed CNN model has a symmetric encoder-decoder architecture with skip connections and is integrated with a FCRF model for post-processing. Atrous convolutions were employed to have large receptive field. Also, transpose convolutions with large filter and overlapping stride were used for upsampling. FCRF model adopted for post-processing accounts for long range dependencies with dense pixel connectivity and refines CNN segmentation outputs. Experimental results on ISPRS Vaihingen dataset are promising. A competitive overall accuracy of 90.5\% was obtained using the proposed model.      


%





\ifCLASSOPTIONcaptionsoff
  \newpage
\fi



%
\bibliography{references}
\bibliographystyle{IEEEtran}
\end{document}